%% file: main.tex
\documentclass[10pt,twocolumn,letterpaper]{article}
\usepackage[pagenumbers]{cvpr} 

\input{preamble}

\definecolor{cvprblue}{rgb}{0.21,0.49,0.74}
\usepackage[pagebackref,breaklinks,colorlinks,citecolor=cvprblue]{hyperref}

\usepackage{graphicx}
\usepackage{amsmath}
\usepackage{amssymb}
\usepackage{booktabs}
\usepackage{makecell}
\usepackage{multirow}
\usepackage{overpic}
\usepackage{comment}
\usepackage{float}
\usepackage[export]{adjustbox}
\usepackage{marvosym}

\def\ie{\emph{i.e.,\ }}

\def\etc{\emph{etc.\ }}

\def\etal{\emph{et al.\ }}

\def\paperID{8694} 
\def\confName{CVPR}
\def\confYear{2024}

\title{A Survey on 3D Skeleton-based Action Recognition Using Learning Method}

\author{
    Bin Ren$^{1,2}$ \quad 
    Mengyuan Liu$^{3}$\textsuperscript{\Letter} \quad 
    Runwei Ding$^4$ \quad 
    Hong Liu$^3$ \\
\thanks{$^{3*}$Equal contribution}
$^1$University of Pisa \quad $^2$University of Trento \quad $^3$Peking University \quad $^4$Peng Cheng Laboratory\\
{\tt\small \Letter~Address correspondence to: nkliuyifang@gmail.com}\\
}

\begin{document}


\twocolumn[{%
\renewcommand\twocolumn[1][]{#1}%
\maketitle
}]

\input{sec/0_abstract}

\input{sec/1_intro}
\input{sec/2_3ds_ar_learning}
\input{sec/3_datasets_performance}

\input{sec/4_discussion}

\input{sec/5_conclusion}

{
    \small
    \bibliographystyle{ieeenat_fullname}
    \bibliography{main}
}

\end{document}

%% file: preamble.tex
\usepackage[dvipsnames]{xcolor}


%% file: sec/0_abstract.tex
\begin{abstract}
3D skeleton-based action recognition (3D SAR) has gained significant attention within the computer vision community, owing to the inherent advantages offered by skeleton data. As a result, a plethora of impressive works, including those based on conventional handcrafted features and learned feature extraction methods, have been conducted over the years. However, prior surveys on action recognition have primarily focused on video or RGB data-dominated approaches, with limited coverage of reviews related to skeleton data. Furthermore, despite the extensive application of deep learning methods in this field, there has been a notable absence of research that provides an introductory or comprehensive review from the perspective of deep learning architectures. To address these limitations, this survey first underscores the importance of action recognition and emphasizes the significance of 3D skeleton data as a valuable modality. Subsequently, we provide a comprehensive introduction to mainstream action recognition techniques based on four fundamental deep architectures, \ie Recurrent Neural Networks (RNNs), Convolutional Neural Networks (CNNs), Graph
Convolutional Network (GCN), and Transformers. All methods with the corresponding architectures are then presented in a data-driven manner with detailed discussion. Finally, we offer insights into the current largest 3D skeleton dataset, NTU-RGB+D, and its new edition, NTU-RGB+D 120, along with an overview of several top-performing algorithms on these datasets. To the best of our knowledge, this research represents the first comprehensive discussion of deep learning-based action recognition using 3D skeleton data.
\end{abstract}

%% file: sec/1_intro.tex
\section{Introduction}
\label{sec:introduction}
\begin{figure}[!t]
    \centering
    \includegraphics[width=1.0\linewidth]{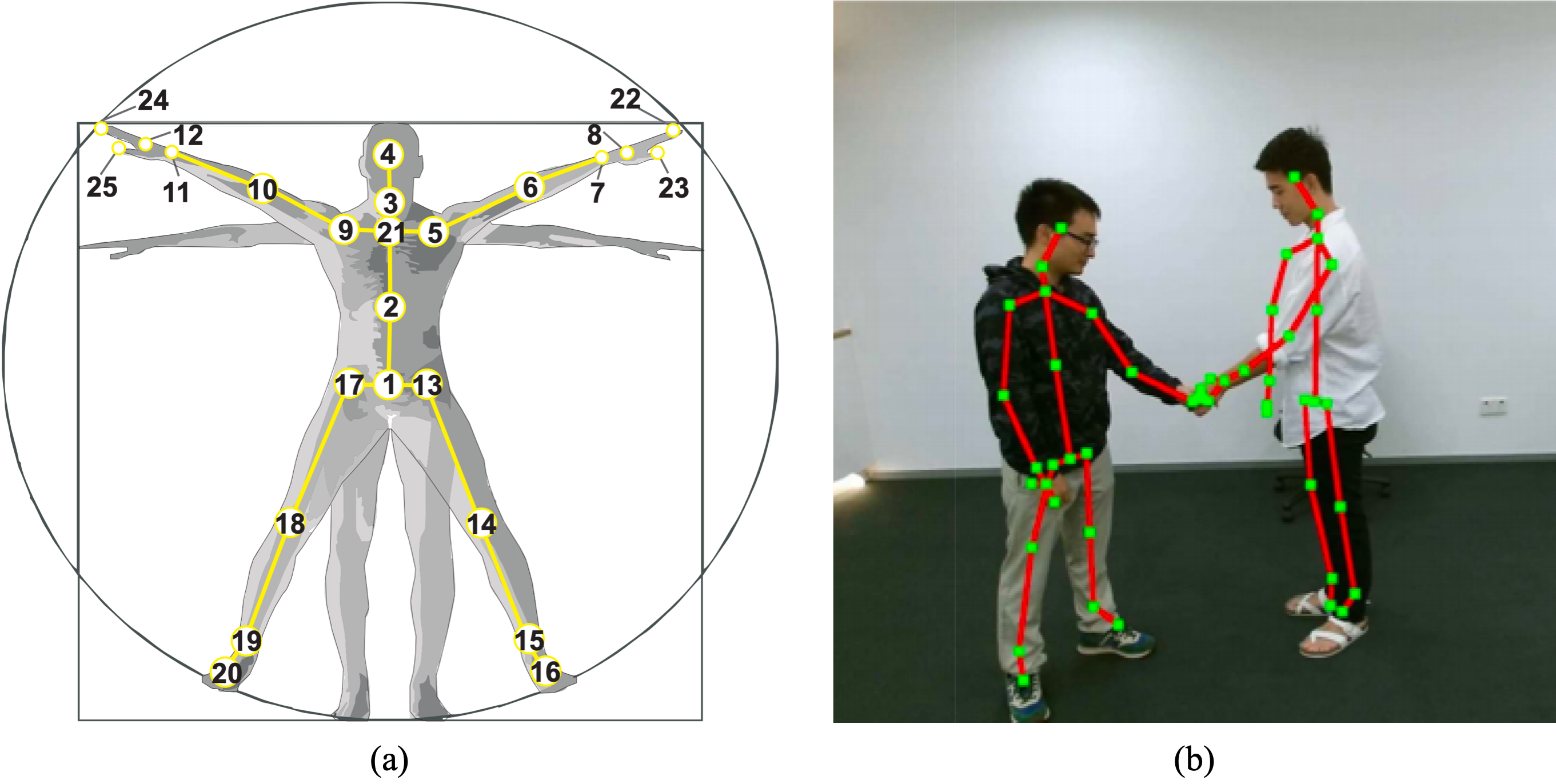}
    \caption{Examples of skeleton data in NTU RGB+D / NTU RGB+D 120 datasets~\cite{shahroudy2016ntu,liu2019ntu}. (a) is the Configuration of 25 body joints in the dataset. (b) illustrates RGB+joints representation of the human body.}
    \label{fig:skeleton_examples}
    \vspace{-4mm}
\end{figure}

Action analysis, a pivotal and vigorously researched topic in the field of computer vision, has been under investigation for several decades \cite{wang2023global,tu2023dtcm,zhang2024facial,wang2024dynamic}. The ability to recognize actions is of paramount importance, as it enables us to understand how humans interact with their surroundings and express their emotions \cite{liu2015sdm, liu2017fusing}. This recognition can be applied across a wide range of domains, including intelligent surveillance systems, human-computer interaction, virtual reality, and robotics \cite{zhang2012imageadmixture, zhang2018detecting, chen2020refinedetlite}. In recent years, the field of skeleton-based action recognition has made significant strides, surpassing conventional hand-crafted methods. This progress has been chiefly driven by substantial advancements in deep learning methodologies.~\cite{ren2011robust,ren2021cloth,liu2017enhanced,yang2019make,liu2017two,tang2023deep,theodoridis2007action,zhao2023modiff,wanghand,liu2023novel}.

Traditionally, action recognition has relied on various data modalities, such as RGB image sequences~\cite{lin2018temporal,feichtenhofer2019slowfast,tran2018closer,liu2020grouped,thatipelli2022spatio}, the depth image sequences~\cite{xu2017lie,baek2016kinematic}, videos, or a fusion of these modalities (e.g., RGB combined with the optical flow)~\cite{simonyan2014two,feichtenhofer2016convolutional,wang2016temporal,gu2013human,hu2015jointly}. These approaches have yielded impressive results through various techniques. Compared to skeleton data, which offers a detailed topological representation of the human body through joints and bones, these alternative modalities often prove computationally intensive and less robust when confronted with complex backgrounds and variable conditions. This includes challenges posed by variations in body scales, viewpoints, and motion speeds~\cite{johansson1973visual,liu2023spatio}.

Furthermore, the availability of sensors like the Microsoft Kinect~\cite{zhang2012microsoft} and advanced human pose estimation algorithms~\cite{chu2017multi,yang2016end,cao2018openpose,zhao2023contextaware} has facilitated the acquisition of accurate 3D skeleton data~\cite{si2019attention}. Figure~\ref{fig:skeleton_examples} provides a visual representation of human skeleton data. In this case, 25 body joints are captured for a given human body. Skeleton sequences possess several advantages over other modalities, characterized by three notable features: 1) Intra-frame spatial information, where strong correlations exist between joints and their adjacent nodes, enabling the extraction of rich structural information. 2) Inter-frame temporal information, which captures strong and clear temporal correlations between frames of each body joint, enhancing the potential for action recognition. 3) A co-occurrence relationship between spatial and temporal domains when considering joints and bones, offering a holistic perspective. These unique attributes have catalyzed substantial research endeavors in human action recognition and detection. The escalating integration of skeleton data is anticipated to pervade diverse applications in the field.

The recognition of human actions using skeleton sequences predominantly hinges on a temporal dimension, transforming it into both a spatial and temporal information modeling challenge. As a result, traditional approaches in skeleton-based methods focus on extracting motion patterns from these sequences, prompting extensive research into handcrafted features.~\cite{vemulapalli2014human,hussein2013human,hu2015jointly,zhou2009hmms,wang2023interweaved,you2023co}. These features often entail capturing the relative 3D rotations and translations among different joints or body parts~\cite{liu2017enhanced,vemulapalli2016rolling}. However, it has become evident that handcrafted features perform well only on specific datasets~\cite{wang2019comparative}, highlighting the challenge that features tailored for one dataset may not be transferable to others. This issue hampers the generalization and broader application of action recognition algorithms.

\begin{figure*}[!t]
	\centering
	\includegraphics[width=1.0\linewidth]{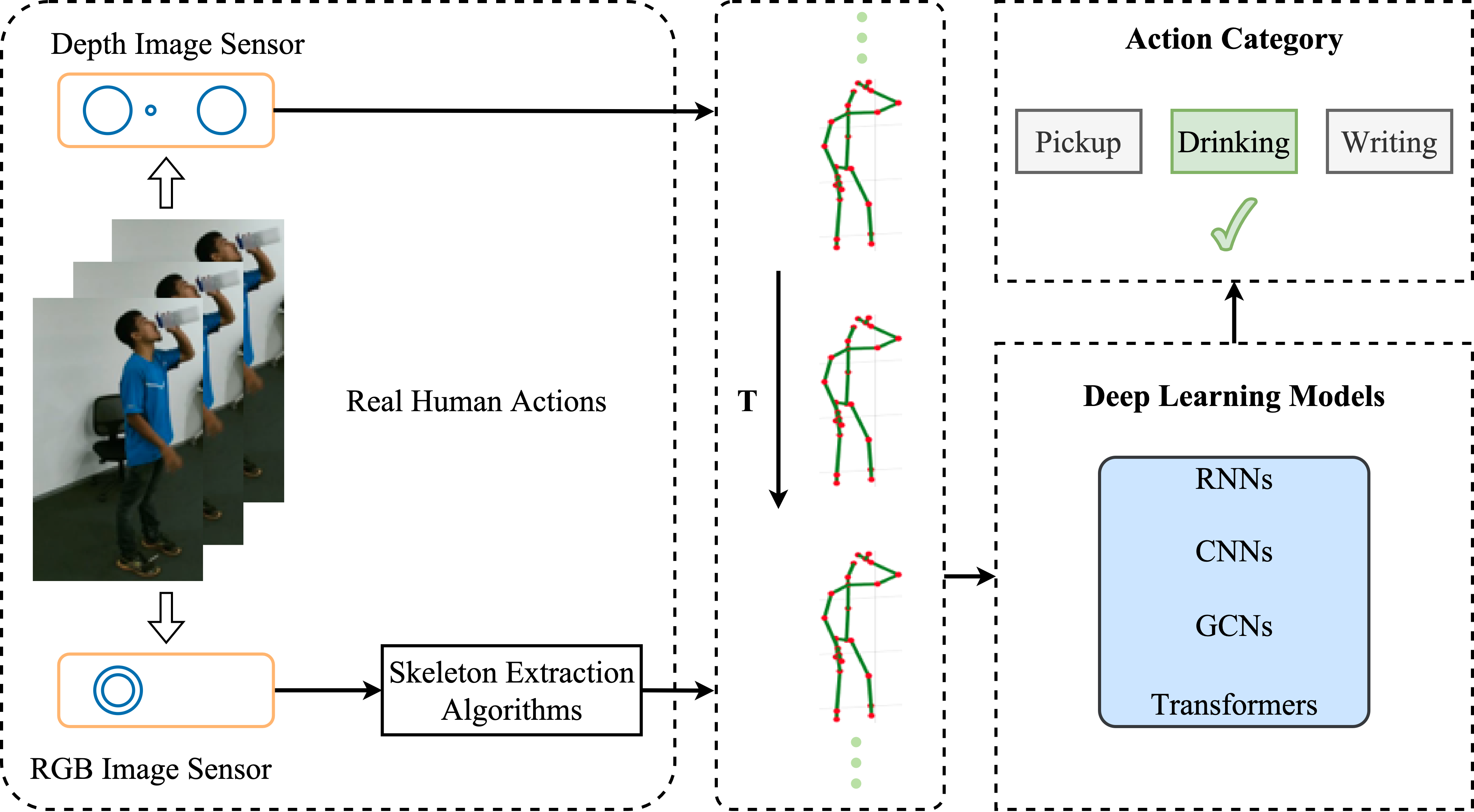}
	\caption{The general pipeline of skeleton-based action recognition using deep learning methods. Firstly, the skeleton data was obtained in two ways, directly from depth sensors or from pose estimation algorithms. The skeleton will be sent into RNNs, CNNs, GCNs, or Transformer-based neural networks. Finally, we get the accurate action category.}
	\label{fig:framework}
\end{figure*}

With the remarkable development and outstanding performance of deep learning methods in various computer vision tasks, such as image classification~\cite{krizhevsky2012imagenet,dosovitskiy2020image} and object detection~\cite{carion2020end,zhu2021deformable}, the application of deep learning to skeleton data for action recognition has gained prominence. Nowadays, deep learning techniques utilizing Recurrent Neural Networks (RNNs)~\cite{lev2016rnn}, Convolutional Neural Networks (CNNs)~\cite{cheron2015p}, Graph
Convolutional Networks (GCNs), and Transformer-based methods have emerged in this field~\cite{yan2018spatial,si2018skeleton}. Figure~\ref{fig:framework} provides an overview of the general pipeline for 3D skeleton-based action recognition (3D SAR) using deep learning, starting from raw RGB sequences or videos and culminating in action category prediction. RNN-based methods leverage skeleton sequences as natural time series data, treating joint coordinates as sequential vectors, aligning well with the RNN's capacity for processing time series information. To enhance the learning of temporal context within skeleton sequences, variants like Long Short-Term Memory (LSTM) and Gated Recurrent Unit (GRU) have been employed. Meanwhile, CNNs complement RNN-based techniques, as they excel at capturing spatial cues in the input data, which RNNs may lack. Additionally, a relatively recent approach, the GCNs has gained attention for its ability to model skeleton data in a natural topological graph structure, with joints and bones as vertices and edges, respectively, offering advantages over alternative formats like images or sequences. \textcolor{black}{Transformer-based methods~\cite{wang20233mformer,zhou2022hypergraph,plizzari2021skeleton,zhu2023stmt,bai2021application,you2023gator} capture the spatial-temporal relation of the input 3D skeleton data mainly based on its core multi-head self-attention mechanism (MSA).}

All these three kinds of deep learning-based architectures have already gained unprecedented performance, but most review works just focus on traditional techniques or deep learning-based methods just with the RGB image or RGB-D data method. Ronald Poppe \emph{et al}.~\cite{poppe2010survey} firstly addressed the basic challenges and characteristics of this domain and then gave a detailed illumination of basic action classification methods about direct classification and temporal state-space models. Daniel and Remi \emph{et al}.~\cite{weinland2011survey} showed an overall overview of the action representation only in both spatial and temporal domains. Though the methods mentioned above provide some inspiration that may be used for input data pre-processing, neither skeleton sequence nor deep learning strategies were taken into account. Recently, Wu \emph{et al}.~\cite{wu2017deep} and Herath \emph{et al}.~\cite{herath2017going} offered a summary of deep learning-based video classification and captioning tasks, in which the fundamental structure of CNN, as well as RNNs, were introduced, and the latter made a clarification about common deep architectures and quantitative analysis for action recognition. To our best knowledge, \cite{Lo3D} is the first work recently giving an in-depth study in 3D SAR, which concludes this issue from the action representation to the classification methods, in the meantime, it also offers some commonly used datasets such as UCF, MHAD, MSR daily activity 3D, \etc \cite{Ellis2013Exploring,Ofli2013Berkeley,Wang2012Mining,liu2022generalized}, while it doesn't cover the emerging GCN based methods. Finally, \cite{wang2019comparative} proposed a new review based on Kinect-dataset-based action recognition algorithms, which organized a thorough comparison of those Kinect-dataset-based techniques with various types of input data including RGB, Depth, RGB+Depth, and skeleton sequences. \cite{sun2022human} presented an overview of the action recognition across all the data modalities but without presenting the Transformer-based methods. In addition, all these works mentioned above also ignore the differences and motivations among CNN-based, RNN-based, GCN-based, and Transformer-based methods, especially when taking the 3D skeleton sequences into account.

To address these issues comprehensively, this survey aims to provide a detailed summary of 3D SAR employing four fundamental deep learning architectures: RNNs, CNNs, GCNs, and Transformers. Additionally, we delve into the motivations behind the choice of these models and offer insights into potential future directions for research in this field.

In summary, our study encompasses four key contributions:
\begin{itemize}
	\item A comprehensive introduction about the superiority of 3D skeleton sequence data and characteristics of three kinds of fundamental deep architectures are presented in a detailed and clear manner, and a general pipeline in 3D SAR using deep learning methods is illustrated. 
	\item Within each type of deep architecture, numerous contemporary methods leveraging skeleton data are introduced, focusing on data-driven approaches. These encompass spatial-temporal modeling, innovative skeleton data representation, and methods for co-occurrence feature learning.
	\item The discussion begins by addressing the latest challenging datasets, notably the NTU-RGB+D 120, along with an exploration of several top-ranked methods. Subsequently, it delves into envisaged future directions in this domain. 
	\item The initial study comprehensively examines four foundational deep architectures, encompassing RNN-based, CNN-based, GCN-based, and Transformer-based methods within the domain of 3D SAR.
\end{itemize}

%% file: sec/2_3ds_ar_learning.tex
\section{3D SAR With Deep Learning}
\label{sec:3d_sar_learning}
While existing surveys have offered comprehensive comparisons of action recognition techniques based on RGB or skeleton data, they often lack a detailed examination from the perspective of neural networks. To bridge this gap, we provide a concise introduction to the fundamental properties of each architecture (Section~\ref{sec:preliminaries}). Then our survey provides an exhaustive discussion and comparison of RNN-based (Section~\ref{sec:rnn_based}), CNN-based (Section~\ref{sec:cnn_based}), GCN-based (Section~\ref{sec:gcn_based}), and Transformer-based (Section~\ref{sec:transformer_based}) methods for 3D skeleton-based action recognition. We will explore these methods in-depth, highlighting their strengths and weaknesses, and introduce several latest related works as case studies, focusing on specific limitations or classic spatial-temporal modeling challenges associated with these neural network models.

\subsection{Preliminaries: Basic Properties of RNNs, CNNs, GCNs, and Transformers}
\label{sec:preliminaries}
\textcolor{black}{Before delving into the specifics of each method, we provide a brief overview of the fundamental architecture, outlining their respective advantages, disadvantages, and coarse selection criteria under the 3D SAR setting.}

\noindent\textbf{\textcolor{black}{RNNs.}} \textcolor{black}{RNNs are ideal for capturing temporal dependencies in sequences of joint movements over time and are suited for modeling action sequences due to their ability to retain temporal information. However, RNNs are also vulnerable to long-term dependencies, potentially missing complex relationships in lengthy sequences, and are computationally inefficient due to sequential processing, leading to longer training times for large-scale datasets.}

\noindent\textbf{\textcolor{black}{CNNs.}} \textcolor{black}{CNNs are not only effective in capturing spatial patterns from the joint coordinates, recognizing spatial features within individual frames of the 3D skeleton data but also great for local spatial relationships among joints. However, CNNs are limited to capturing temporal evolution in sequences, potentially missing out on the temporal dynamics crucial for action recognition.}
 
\noindent\textbf{\textcolor{black}{GCNs.}} \textcolor{black}{GCNs are designed to manage graph-structured data such as skeletal joint connections in action recognition, enabling the learning of relationships between joints and their connectivity while integrating spatial and temporal information. However, GCNs can be sensitive to noisy or irregular connections among joints, potentially impacting recognition accuracy, particularly in complex actions.}

\noindent\textbf{\textcolor{black}{Transformers.}} \textcolor{black}{Transformers is not only efficient at capturing long-range dependencies without the vanishing/exploding gradient issue but also Versatile in handling multiple modalities and learning global relationships. However, it's also computationally intensive due to attention mechanisms, potentially requiring substantial computational resources. What's more, compared to RNNs, it is also limited to sequential locality}

\begin{figure*}[!t]
	\hfill
	\begin{minipage}[b]{1.0\linewidth}
		\centering
		\centerline{\includegraphics[width=0.9\linewidth]{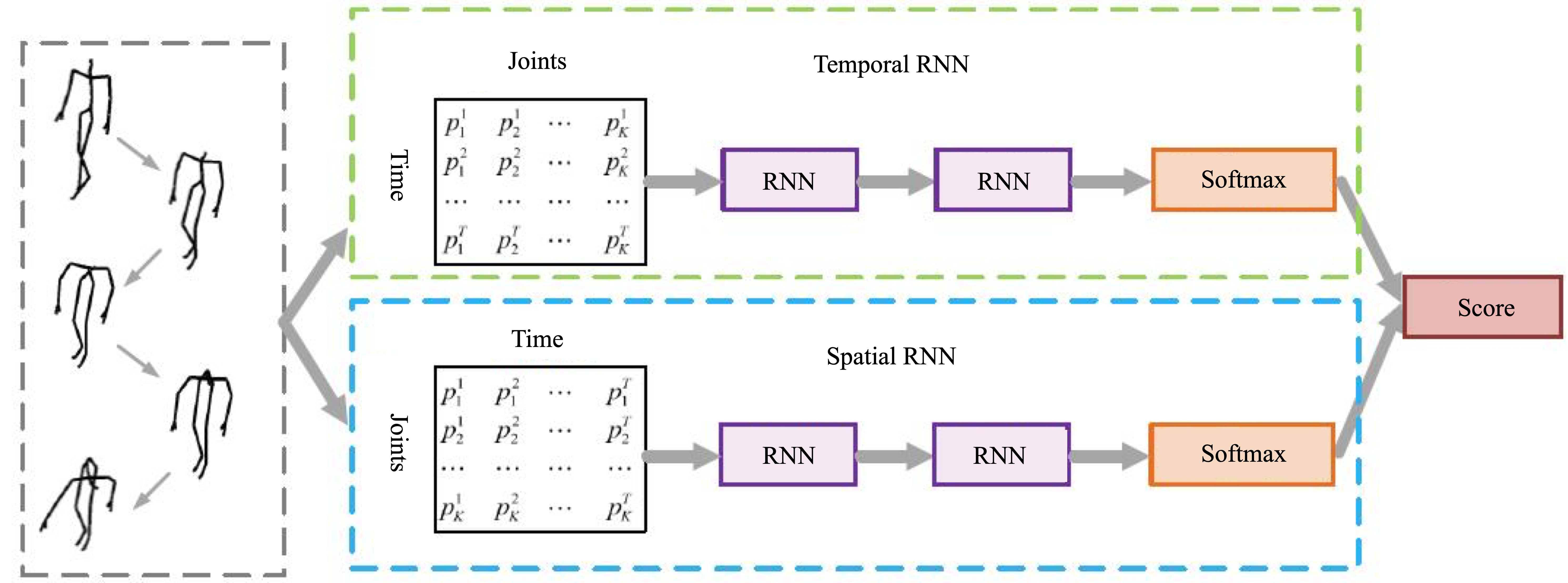}}
		\vspace{-0.1cm}
		\centerline{(a)}\medskip
	\end{minipage}
	\hfill
	\begin{minipage}[b]{1.0\linewidth}
		\centering
		\centerline{\includegraphics[width=0.95\linewidth]{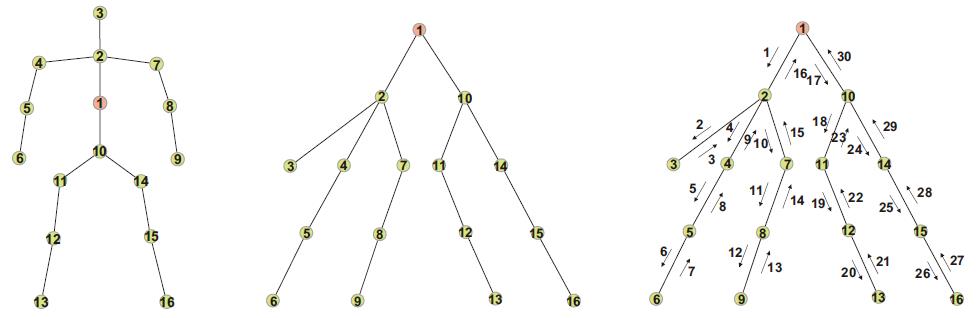}}
		\vspace{-0.1cm}
		\centerline{(b)}\medskip
	\end{minipage}
	\vspace{-0.5cm}
	\caption{Examples of mentioned methods for dealing with spatial modeling problems. (a) shows a two-stream framework that enhances the spatial information by adding a new stream~\cite{wang2017modeling}. (b) illustrates a data-driven technique that addresses the spatial modeling ability by giving a transform towards original skeleton sequence data~\cite{liu2016spatio}.}
	\label{fig:fig3}
	\vspace{-0.4cm}
\end{figure*}

\subsection{RNN-Based Methods}
\label{sec:rnn_based}
Recursive connections within the RNN structure are established by feeding the output of the previous time step as the input to the current time step, as demonstrated in prior work~\cite{zhang2018adding}. This approach is known to be effective for processing sequential data. In a similar vein, models like the standard RNN, LSTM, and GRU were introduced to address limitations such as gradient-related issues and the modeling of long-term temporal dependencies that were present in the standard RNN.

From the first aspect, spatial-temporal modeling can be seen as the principle in action recognition tasks. Due to the weakness of the spatial modeling ability of RNN-based architecture, the performance of some related methods generally could not gain a competitive result~\cite{wu2014leveraging,zhao2017two,li2017adaptive}. Recently, Hong \emph{et al}.~\cite{wang2017modeling} proposed a novel two-stream RNN architecture to model both temporal dynamics and spatial configurations for skeleton data. Figure~\ref{fig:fig3} shows the framework of their work. An exchange of the skeleton axes was applied for the data level pre-processing for the spatial dominant learning. Unlike~\cite{wang2017modeling}, Jun \emph{et al}.~\cite{liu2016spatio} stepped into the traversal method of a given skeleton sequence to acquire the hidden relationship of both domains. Compared with the general method which arranges joints in a simple chain so that ignores the kinetic dependency relations between adjacent joints, the mentioned tree-structure-based traversal would not add false connections between body joints when their relation is not strong enough. Then, using an LSTM with a trusted gate treat the input discriminately, through which, if the tree-structured input unit is reliable, the memory cell will be updated by importing input latent spatial information. Inspirited by the property of CNN, which is extremely suitable for spatial modeling. 
Li \emph{et al}.\cite{li2021memory} incorporated an attention RNN with a CNN model to enhance the complexity of spatial-temporal modeling. Initially, they introduced a temporal attention module within a residual learning module, allowing for the recalibration of temporal attention across frames within a skeleton sequence. Subsequently, they applied a spatial-temporal convolutional module to this first module, treating the calibrated joint sequences as images. Furthermore, in the work by Lin \emph{et al}.\cite{li2018skeleton}, an attention recurrent relation LSTM network was employed. This network combines a recurrent relation network for spatial features with a multi-layer LSTM to capture temporal features within skeleton sequences.

The second aspect involves the network structure, serving as a solution to address the limitations of standard RNNs. While RNNs are inherently suitable for sequence data, they often suffer from well-known problems like gradient exploding and vanishing. Although LSTM and GRU have alleviated these issues to some extent, the use of hyperbolic tangent and sigmoid activation functions can still result in gradient decay across layers. In response, new types of RNN architectures have been proposed~\cite{bradbury2016quasi,lei2018training,li2018independently}. Shuai \emph{et al}.~\cite{li2018independently} introduced an independently recurrent neural network (IndRNN) designed to address gradient exploding and vanishing problems, making it feasible and more robust to construct longer and deeper RNNs for high-level semantic feature learning. This modification for RNNs is not limited to skeleton-based action recognition but can also find applications in other domains, such as language modeling. In the IndRNN structure, neurons in one layer operate independently of each other, enabling the processing of much longer sequences.

\begin{figure*}[!t]
\hfill
\begin{minipage}[b]{1.0\linewidth}
	\centering
	\centerline{\includegraphics[width=0.9\linewidth]{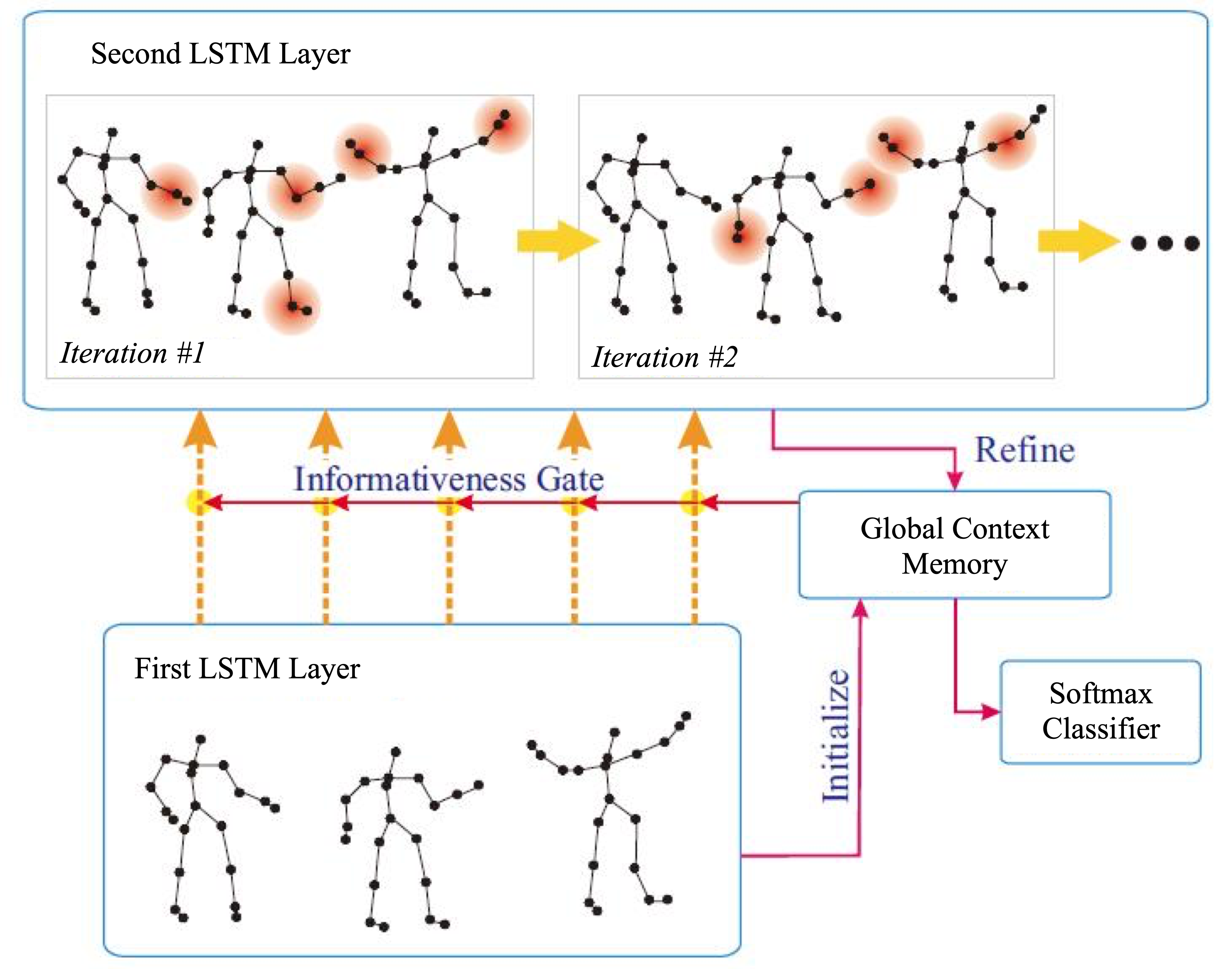}}
	\vspace{-0.1cm}
	\centerline{(a)}\medskip
\end{minipage}
\hfill
\begin{minipage}[b]{1.0\linewidth}
	\centering
	\centerline{\includegraphics[width=0.9\linewidth]{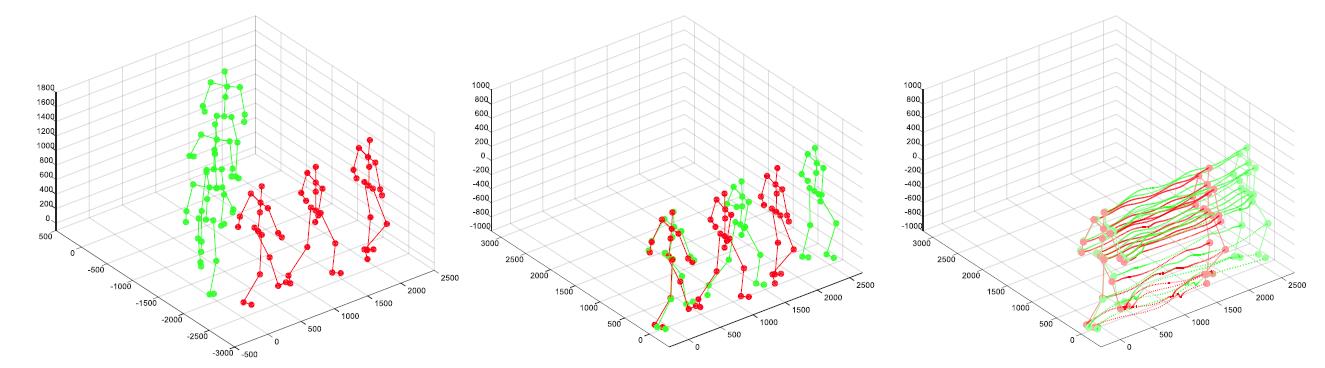}}
	\vspace{-0.1cm}
	\centerline{(b)}\medskip
\end{minipage}
\vspace{-0.5cm}
\caption{Data-driven based method. (a) shows the different importance among different joints for a given skeleton action~\cite{liu2017global}. (b) give shows the feature representation processes, from left to right are original input skeleton frames, transformed input frames, and extracted salient motion features respectively~\cite{lee2017ensemble}.}
\label{fig:fig4}
\vspace{-0.4cm}
\end{figure*}

Finally, the third aspect is the data-driven. In the consideration that not all joints are informative for an action analysis, \cite{liu2017global} add global context-aware attention to LSTM networks, which selectively focus on the informative joints in a skeleton sequence. Figure~\ref{fig:fig4} illustrates the visualization of the proposed method, from the figure we can conclude that the more informative joints are addressed with a red circle color area, indicating those joints are more important for this special action. In addition, because the skeletons provided by datasets or depth sensors are not perfect, which would affect the result of an action recognition task, \cite{lee2017ensemble} transform skeletons into another coordinate system for the robustness to scale, rotation and translation first and then extract salient motion features from the transformed data instead of sending the raw skeleton data to LSTM. Figure~\ref{fig:fig4}(b) shows the feature representation process.

Numerous valuable works have utilized RNN-based methods to address challenges related to large viewpoint changes and the relationships among joints within a single skeleton frame. However, it's essential to acknowledge that in specific modeling aspects, RNN-based methods may exhibit limitations compared to CNN-based approaches. In the following sections, we delve into an intriguing question: how do CNN-based methods perform temporal modeling, and how can they strike the right balance between spatial and temporal information in action recognition?

\begin{figure*}[!t]
	\hfill
	\begin{minipage}[b]{1.0\linewidth}
		\centering
		\centerline{\includegraphics[width=1.0\linewidth]{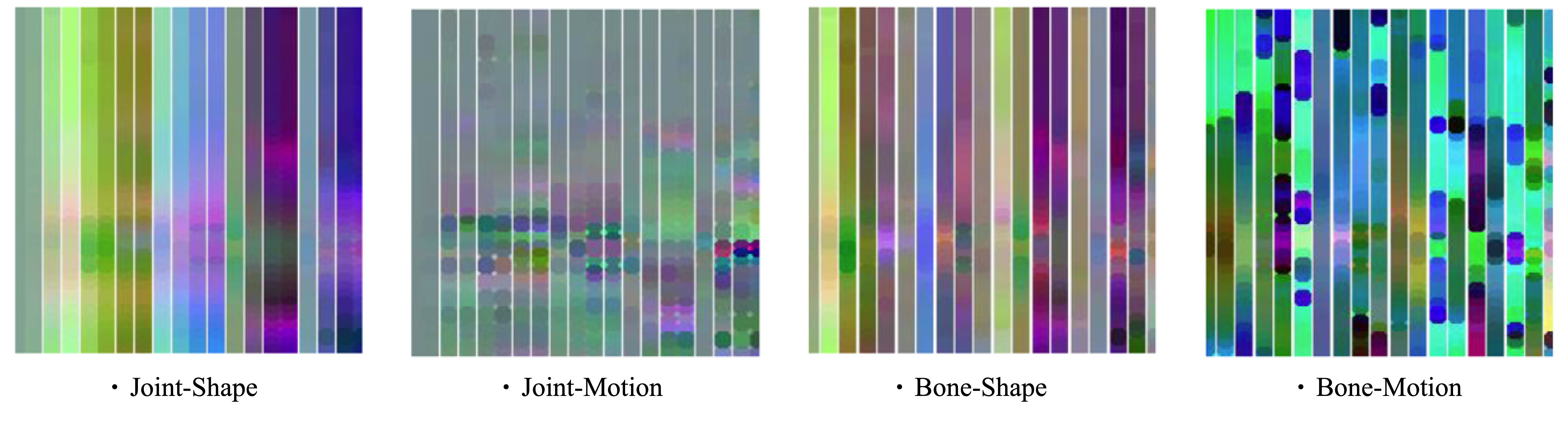}}
		\vspace{-0.1cm}
		\centerline{(a)}\medskip
	\end{minipage}
	\hfill
	\begin{minipage}[b]{1.0\linewidth}
		\centering
		\centerline{\includegraphics[width=1.0\linewidth]{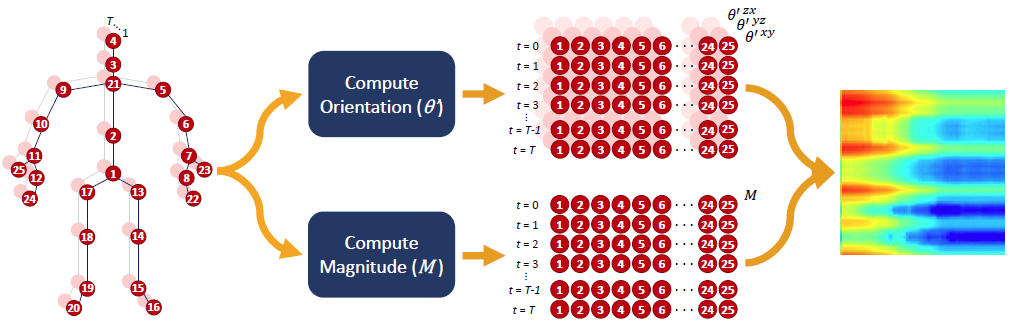}}
		\vspace{-0.1cm}
		\centerline{(b)}\medskip
	\end{minipage}
	\vspace{-0.5cm}
	\caption{Examples of the proposed representation skeleton image. (a) shows Skeleton sequence shape-motion representations~\cite{li2019learning} generated from "pick up with one hand" on Northwestern-UCLA dataset\cite{wang2014cross} (b) shows the SkeleMotion representation workflow~\cite{caetano2019skeleton}.}
	\label{fig:fig5}
	\vspace{-0.4cm}
\end{figure*}

\subsection{CNN-Based Methods}
\label{sec:cnn_based}
While convolutional neural networks (CNNs) offer efficient and effective high-level semantic cue learning, they are primarily tailored for regular image tasks. However, action recognition from skeleton sequences presents a distinct challenge due to its inherent time-dependent nature. Achieving the right balance and maximizing the utilization of both spatial and temporal information within a CNN-based architecture remains a challenging endeavor.

Typically, from the spatial-temporal modeling aspect, most of the CNN-based methods explored the representation of 3D skeleton sequences, Specifically, to accommodate the input requirements of CNNs, 3D-skeleton sequence data undergoes the transformation from a vector sequence to a pseudo-image format. However, achieving a suitable representation that effectively combines both spatial and temporal information can be challenging. 
Consequently, many researchers opt to encode skeleton joints into multiple 2D pseudo-images, which are subsequently fed into CNNs to facilitate the learning of informative features~\cite{ding2017investigation,xu2018ensemble}. Wang \emph{et al}.~\cite{Wang2016Action} proposed the Joint Trajectory Maps (JTM), which represent spatial configuration and dynamics of joint trajectories into three texture images through color encoding. However, this kind of method is a little complicated and also loses importance during the mapping procedure. 
To tackle this shortcoming, Li \emph{et al}.~\cite{Bo2017Skeleton} used a translation-scale invariant image mapping strategy which firstly divided human skeleton joints in each frame into five main parts according to the human physical structure, then those parts were mapped to 2D form. This method makes the skeleton image consist of both temporal information and spatial information. However, though the performance was improved, there is no reason to take skeleton joints as isolated points, cause in the real world, imitate connection exists among our body, for example, when waiting for the hands, not only the joints directly within the hand should be taken into account, but also other parts such as shoulders and legs are considerable. 
Li \emph{et al}.~\cite{li2019learning} proposed the shape-motion representation from geometric algebra, which addressed the importance of both joints and bones and fully utilized the information provided by the skeleton sequence. Similarly, \cite{liu2017enhanced} also use the enhanced skeleton visualization to represent the skeleton data, and Carlos \etal \cite{caetano2019skelemotion} also proposed a new representation named SkeleMotion based on motion information that encodes the temporal dynamics by explicitly computing the magnitude and orientation values of the skeleton joints. Figure~\ref{fig:fig5} (a) shows the shape-motion representation proposed by~\cite{li2019learning} while Figure~\ref{fig:fig5} (b) illustrate the SkeleMotion representation. What's more, similarly to SkeleMotion, \cite{caetano2019skeleton} uses the framework of SkeleMotion but is based on tree structure and reference joints for a skeleton image representation.

Commonly, CNN-based methods represent a skeleton sequence as an image by encoding temporal dynamics and skeleton joints as rows and columns, respectively. 
However, this simplistic approach may limit the model's ability to capture co-occurrence features, as it considers only neighboring joints within the convolutional kernel and may overlook latent correlations involving all joints. Consequently, CNNs might fail to learn the corresponding and useful features. In response to this limitation, Chao \emph{et al}.~\cite{Chao2018Co} introduced an end-to-end framework designed to learn co-occurrence features through a hierarchical methodology. This approach gradually aggregates different levels of contextual information, beginning with the independent encoding of point-level information, which is then assembled into semantic representations within both temporal and spatial domains.

\textcolor{black}{Besides explorations in the representation of 3D skeleton sequences, there also exist some other problems in CNN-based techniques. For example, to find a balance between the model size and the corresponding inference efficiency, DD-Net~\cite{yang2019make} was proposed to model double feature and double motion via CNN for efficient solutions. Kim \etal~\cite{soo2017interpretable} proposed to use the temporal CNN (TCN) for modeling the interpretable spatio-temporal cues~\cite{lea2017temporal}. As a result, the point-level feature of each joint is learned. In addition, two-stream and three-stream CNN-based heavy models are also proposed for improving the 
representation learning ability for spatial-temporal modeling~\cite{Ruiz20173D}. So the skeleton-based action recognition using CNN is still an open problem waiting for researchers to dig in.}

\subsection{GCN-Based Methods}
\label{sec:gcn_based}
Drawing inspiration from the inherent topological graph structure of human 3D-skeleton data, distinct from the sequential vector or pseudo-image treatments in RNN-based or CNN-based methods. Recently the Graph Convolution Network has been adopted in this task frequently due to the effective representation of the graph structure data. Generally, two kinds of graph-related neural networks can be found, \ie the graph and recurrent neural networks, and graph and convolutional neural networks (GCNs). In this survey, we mainly pay attention to the latter. This focus yielded compelling results, as evidenced by the performance of the GCN-based method on the rank board. Furthermore, merely encoding the skeleton sequence into a vector or 2D grid fails to fully capture the interdependence among correlated joints from the skeleton's perspective. 
Conversely, GCNs present adaptability to diverse structures, such as the skeleton graph. Nonetheless, the principal challenge within GCN-based approaches persists in the handling of skeleton data, particularly in structuring the original data into a coherent graph format. Yan \etal \cite{yan2018spatial} first presented a novel model, the spatial-temporal graph convolutional networks (ST-GCN), for skeleton-based action recognition. Specifically, the approach first involved the creation of a spatial-temporal graph, wherein the joints functioned as graph vertices, establishing inherent connections within the human body structure and across temporal sequences as the graph edges. Following this step, the ST-GCN's higher-level feature maps on the graph underwent classification using a standard Softmax classifier, assigning them to their respective action categories. 
This work has notably directed attention towards employing GCNs for skeleton-based action recognition, resulting in a surge of recent related research~\cite{shi2019two,zhang2020semantics,cheng2020decoupling,chi2022infogcn,duan2022revisiting,zhou2023learning}. 

\begin{figure}[t]
	\centering
	\includegraphics[width=1.0\linewidth]{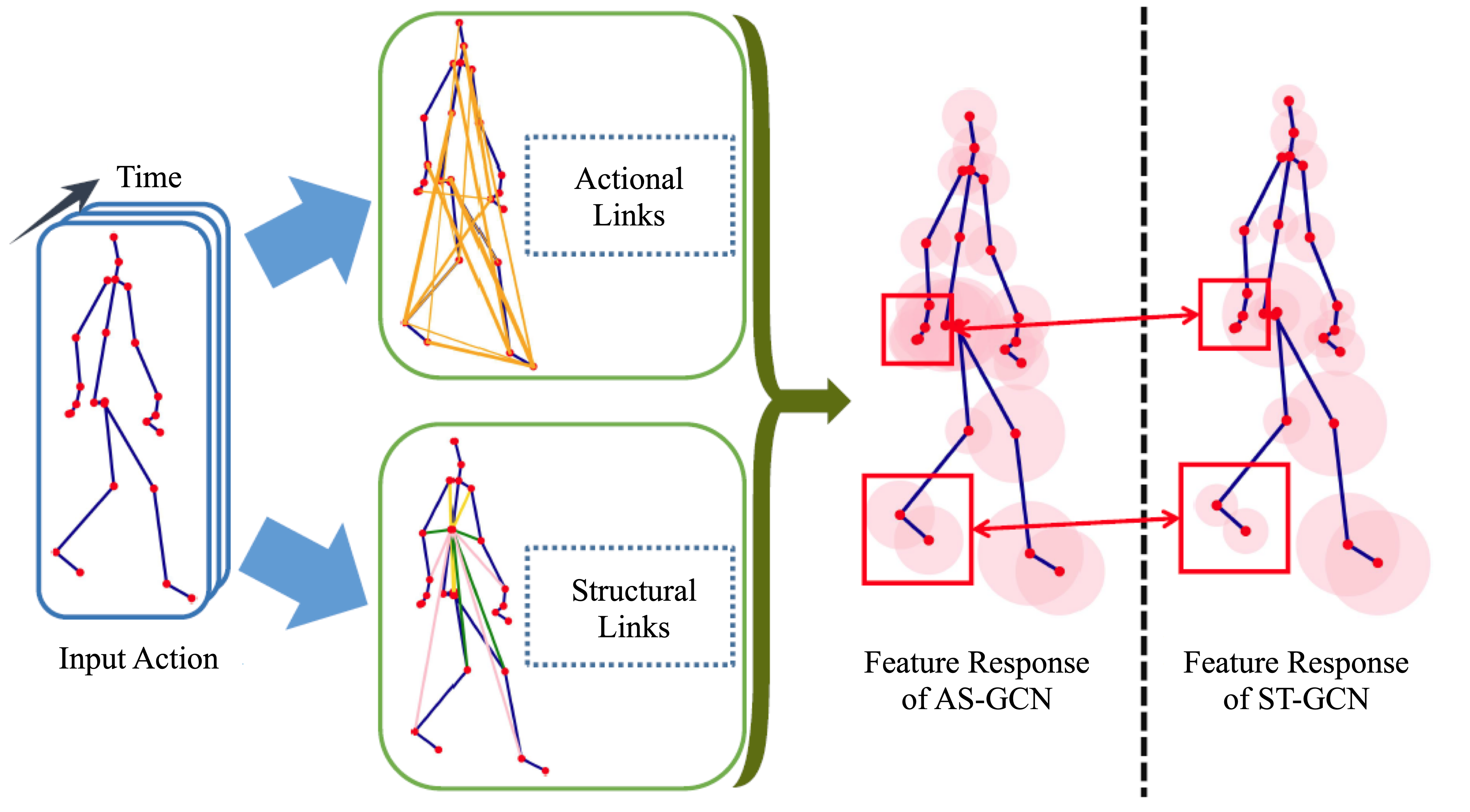}
	\caption{Feature learning with generalized skeleton graphs~\cite{li2019actional}.}
	\label{fig:fig6}
	\vspace{-0.4cm}
\end{figure}

\textcolor{black}{Built upon GCNs, two main common aspects are explored, \ie more representative manner for the construction of the skeleton data and more effective designs of the GCN-based model~\cite{2sagcn2019cvpr,li2019actional}.} 

\textcolor{black}{From the first aspect, \cite{li2019actional} proposed the Action-Structural Graph Convolutional Networks (AS-GCN) could not only recognize a person's action but also use a multi-task learning strategy to output a prediction of the subject's next possible pose. The constructed graph in this work can capture richer dependencies among joints by two modules called Actional Links and Structural Links. Figure~\ref{fig:fig6} shows the feature learning and its generalized skeleton graph of AS-GCN. Multi-task learning strategy used in this work may be a promising direction because the target task would be improved by the other task as a complementary. To capture and enhance richer feature representations, Shi \etal \cite{shi2019two} introduced the 2s-AGCN, which incorporates an adaptive topology graph. This approach allows for automatic updates leveraging the neural network's backpropagation algorithm, effectively enhancing the characterization of joint connection strengths. Liu \etal \cite{liu2020disentangling} proposes MS-G3D which constructs a unified spatial-temporal graph. This big spatial-temporal graph is composed of several subgraphs, and each subgraph represents the spatial
relationships of joints on a certain frame. This form of the adjacent matrix can effectively model the relationship between different joints in different frames. Similarly, there are also a lot of following-up methods proposed for constructing more representative graphs~\cite{wang2022skeleton,hao2021hypergraph,lee2023hierarchically}.}

\textcolor{black}{From the second aspect, traditional GCNs operate as straight feed-forward networks, limiting low-level layers' access to semantic information from higher-level layers. To address this, Yang \etal \cite{yang2021feedback} introduce the Feedback Graph Convolutional Network (FGCN) aimed at incrementally acquiring global spatial-temporal features. Departing from direct manipulation of the complete skeleton sequence, FGCN adopts a multi-stage temporal sampling strategy to sparsely extract a sequence of input clips from the skeleton data. Furthermore, Bian \etal \cite{bian2021structural} introduces a structural knowledge distillation scheme aimed at mitigating accuracy loss resulting from low-quality data, thereby enhancing the model's resilience to incomplete skeleton sequences. Fang \etal \cite{fang2022spatial} presents the spatial-temporal slow-fast graph convolutional network (STSF-GCN), which conceptualizes skeleton data akin to a unified spatial-temporal topology, reminiscent of MS-G3D.}

\textcolor{black}{From the preceding introduction and discussion, it's evident that the predominant concern revolves around data-driven approaches, seeking to uncover latent insights within 3D skeleton sequence data. In the realm of GCN-based action recognition, the central query persists: 'How do we extract this latent information?' This question remains an ongoing challenge. Particularly noteworthy is the inherent temporal-spatial correlation within the skeleton data itself. The optimal utilization of these two aspects warrants further exploration. There remains substantial potential for enhancing their effectiveness, calling for deeper investigation and innovative strategies to maximize their utilization.}

\subsection{\textcolor{black}{Transformer-Based Methods}}
\label{sec:transformer_based}
\textcolor{black}{Transformers~\cite{vaswani2017attention} demonstrated their overwhelming power on a broad range of language tasks (\emph{e.g.}, text classification, machine translation, or question answering~\cite{vaswani2017attention,khan2022transformers}), and the vision community
follows it closely and extends it for vision tasks, such as image classification~\cite{dosovitskiy2020image,touvron2021training,ren2023masked}, object detection~\cite{carion2020end,zhu2021deformable}, segmentation~\cite{ye2019cross}, image restoration~\cite{chen2021pre,li2023efficient}, and point cloud registration~\cite{mei2023overlap,huang2023cross,wang2023zero}. The emergence of transformer algorithms marks a pivotal shift in point-centric research. These transformer-based methods are gradually challenging the dominance of GCN methods, showcasing promising advancements in computational efficiency and accuracy. Upon analysis, we firmly believe that transformer-based approaches retain robust potential and are poised to become the mainstream technique in the future.}

\textcolor{black}{The core module in Transformer, multi-head self-attentions (MSAs)~\cite{vaswani2017attention,dosovitskiy2020image} aggregate sequential tokens with normalized attention as:
$\pmb{z}_j = \sum_i \texttt{Softmax}(\frac{\pmb{Q}\pmb{K}}{\sqrt{d}})_i\pmb{V}_{i,j}$
where $\pmb{Q}$, $\pmb{K}$ and $\pmb{V}$ are query, key and value matrices, respectively. $d$ is the dimension of query and key, and $\pmb{z}_j$ is the $j$-th output token. This step usually represents the context relation computation and update of the overall 3D skeleton features. Building upon the MSA from the Transformer for solving the 3D-SAR problem, there are lots of transformer architecture-based solutions are proposed}

In particular, Cho \etal \cite{cho2020self} proposed a novel model called Self-Attention Network (SAN) that completely utilizes the self-attention mechanism to model spatial-temporal correlations. Shi \etal \cite{shi2020decoupled} proposed a decoupled spatial-temporal attention network (DSTA-Net) that contains spatial-temporal attention decoupling, decoupled position encoding, and global spatial regularization. DSTA-Net decouples the skeleton data into four streams, namely, spatial-temporal stream, spatial stream, slow-temporal stream, and fast-temporal stream, each data stream focuses on expressing a particular aspect of the action. Plizzari \etal \cite{plizzari2021spatial} proposed a novel Spatial-Temporal Transformer network (ST-TR) in which the spatial self-attention module and temporal self-attention module are used to capture the correlation between different nodes in a frame and the dynamic relationship between the same node in the whole frames. To handle action sequences of varying lengths proficiently, Ibh \etal \cite{ibh2023tempose} proposed TemPose which leaves out the padded temporal and interaction tokens in the attention map. At the same time, Tempose codes the position of the player and the position of the badminton ball to predict the action class together.

\input{tables/sota_comparison}

The Transformer-based approach effectively mitigates the issue of solely concentrating on local information and excels in capturing extensive dependencies over long sequences. When applied to tasks involving skeleton-based human behavior recognition, the Transformer architecture demonstrates adeptness in capturing temporal relationships. However, its efficacy in modeling spatial relationships remains constrained due to limitations in capturing and encoding the intricate high-dimensional semantic information inherent in skeleton data~\cite{zhu2023motionbert,xiang2023generative}. Simultaneously, numerous approaches have emerged that amalgamate the Transformer with GCNs or CNNs, thereby forming hybrid architectures. These models are designed with the aspiration of harnessing the strengths inherent in each fundamental architecture. By combining the Transformer's capabilities with the specialized strengths of RNNs, CNNs, or GCNs, these hybrid models aim to achieve a more comprehensive and powerful framework for diverse tasks~\cite{yuan2022spatial,zhou2022hypergraph,zhang2022zoom,gao2022focal}.

%% file: tables/sota_comparison.tex
\begin{table*}[!t]
	\centering
	\caption{\textcolor{black}{The Performance of the latest state-of-the-art 3D Skeleton-based methods on NTU-RGB+D dataset.}}
	\label{tab:table1}
        \resizebox{1.\textwidth}{!}{
	\begin{tabular}{cccccc}
		\toprule[1pt]
		\multicolumn{6}{c}{\textbf{NTU-RGB+D dataset}}                            \\ 
		Rank & Paper & Year & Accuracy (C-View) & Accuracy (C-Subject) & Method \\ \Xhline{0.8pt}
            1     &Wang \emph{et al}.~\cite{wang20233mformer} & 2023 & 98.7 & 94.8 & Two-stream Transformer \\ 
            2 & Duan \etal ~\cite{duan2022dg} & 2022 & 97.5 & 93.2 & Dynamic group GCN \\ 
            3     &Liu \emph{et al}.~\cite{liu2023temporal} & 2023 & 96.8 & 92.8 & Temporal decoupling GCN\\ 
            4 & Zhou \etal ~\cite{zhou2022hypergraph} & 2022 & 96.5 & 92.9 & Transformer \\ 
            
            5 & Chen \etal ~\cite{chen2021channel} & 2021 & 96.8 & 92.4 & Topology refinement GCN \\ 
            
            6 & Zeng \etal ~\cite{zeng2021learning} & 2021 & 96.7 & 91.6 & Skeletal GCN \\ 

            7 & Liu \etal ~\cite{liu2020disentangling} & 2020 & 96.2 & 91.5 & Disentangling and unifying GCN \\ 

            8 & Ye \etal ~\cite{ye2020dynamic} & 2020 & 96.0 & 91.5 & Dynamic GCN \\ 

		9 & Shi \etal~\cite{shi2019skeleton}        & 2019 &96.1 &89.9 & Directed graph neural networks\\ 
		10 &Shi \emph{et al}.~\cite{shi2019two}             & 2018 &95.1 &88.5 &Two-stream adaptive GCN\\ 
		11    &Zhang \emph{et al}.~\cite{zhang2019view}        & 2018 &95.0 &89.2 &LSTM based RNN        \\ 
		12    &Si \emph{et al}.~\cite{si19atten}               & 2019 &95.0 &89.2 &AGC-LSTM(Joints\&Part)        \\ 
		13    &Hu \emph{et al}.~\cite{hu2019skeleton}          & 2018 &94.9 &89.1 &Non-local S-T + frequency attention    \\ 
		14     &Li \emph{et al}.~\cite{li2019actional}            & 2019 &94.2 &86.8 &GCN        \\ 
		
		15     &Liang \emph{et al}.~\cite{liang2019three}       & 2019 &93.7 &88.6 &3S-CNN + multi-task ensemble learning    \\ 
		
		16     &Song \emph{et al}.~\cite{song2019richlt}        & 2019 &93.5 &85.9 &Richly activated GCN     \\
		17     &Zhang \emph{et al}.~\cite{zhang2019semantics}   & 2019 &93.4 &86.6 &Semantics-guided GCN     \\ 
		18    &Xie \emph{et al}.~\cite{li2021memory}          & 2018 &93.2 &82.7 &RNN+CNN+Attention    \\ \bottomrule[0.1em]     
	\end{tabular}
        }
\end{table*}

\begin{table*}[t]
	\centering
	\caption{\textcolor{black}{The Performance of the latest state-of-the-art 3D Skeleton-based methods on NTU-RGB+D 120 dataset.}}
	\label{tab:table2}
        \resizebox{1.\textwidth}{!}{
	\begin{tabular}{cccccc}
		\toprule[0.1em]    
		\multicolumn{6}{c}{\textbf{NTU-RGB+D 120 dataset}}                        \\ 
		Rank & Paper & Year & Accuracy (C-Subject) & Accuracy (C-Setup) & Method \\ 
		\Xhline{0.8pt}
            1     &Wang \emph{et al}.~\cite{wang20233mformer} & 2023 & 92.0 & 93.8 & Two-stream Transformer \\ 
            2     &Xu \emph{et al}.~\cite{xu2023language} & 2023 & 90.7 & 91.8 & Language Knowledge-Assisted \\

            3     &Zhou \etal \cite{zhou2022hypergraph} & 2022 & 89.9 & 91.3 & Transformer \\ 

            4     &Duan \etal \cite{duan2022dg} & 2022 & 89.6 & 91.3 & Dynamic group GCN \\ 
            
            5     &Chen \etal \cite{chen2021channel} & 2021 & 88.9 & 90.6 & Topology refinement GCN \\ 

            6     &Chen \etal \cite{chen2021learning} & 2021 & 88.2 & 89.3 & Spatial-Temporal GCN \\ 

            7     &Liu \etal \cite{liu2020disentangling} & 2020 & 86.9 & 88.4 & Disentangling and unifying GCN \\ 

            8     &Cheng \etal \cite{cheng2020skeleton} & 2020 & 85.9 & 87.6 & Shift GCN \\

		9     &Caetano \emph{et al}.~\cite{caetano2019skeleton}& 2019 &67.9 &62.8 &Tree Structure + CNN    \\ 
		
		10     &Caetano \emph{et al}.~\cite{caetano2019skelemotion}  & 2019 &67.7 &66.9 &SkeleMotion        \\ 
		
		11    &Liu \emph{et al}.~\cite{liu2018recognizing}     & 2018 &64.6 &66.9 &Body Pose Evolution Map        \\ 
		
		12     &Ke \emph{et al}.~\cite{ke2018learning}          & 2018 &62.2 &61.8 &Multi-Task CNN with RotClips        \\ 
		
	      13   &Liu \emph{et al}.~\cite{liu2017skeleton}        & 2017 &61.2 &63.3 &Two-Stream Attention LSTM        \\ 
		
		14     &Liu \emph{et al}.~\cite{liu2017enhanced}        & 2017 &60.3 &63.2 &Skeleton Visualization (Single Stream)        \\ 
		
		15     &Jun \emph{et al}.~\cite{liu2019skeleton}        & 2019 &59.9 &62.4 &Online+Dilated CNN         \\ 
		
		16     &Ke \emph{et al}.~\cite{ke2017new}               & 2017 &58.4 &57.9 &Multi-Task Learning CNN                                         \\ 
		
		17     &Jun \emph{et al}.~\cite{liu2017global}          & 2017 &58.3 &59.2 &Global Context-Aware Attention LSTM        \\ 
		
		18    &Jun \emph{et al}.~\cite{liu2016spatio}          & 2016 &55.7 &57.9 &Spatio-Temporal LSTM        \\ 
		\bottomrule[0.1em]
	\end{tabular}
        }
\end{table*}

%% file: sec/3_datasets_performance.tex
\section{Latest Datasets And Performance}
\label{sec:datasets_performance}
Skeleton sequence datases such as MSRAAction3D~\cite{Li2010Action}, 3D Action Pairs~\cite{Oreifej2013HON4D}, and MSR Daily Activity3D~\cite{Wang2012Mining} have be analysed in lots of previous surveys~\cite{Lo3D,herath2017going,wang2019comparative}. In this survey, we mainly address the following two recent datasets, NTU-RGB+D~\cite{shahroudy2016ntu} and NTU-RGB+D 120~\cite{liu2019ntu}.

The NTU-RBG+D dataset, introduced in 2016, stands as a significant resource, comprising 56,880 video samples gathered through Microsoft Kinect-v2. This dataset holds a prominent position as one of the largest collections available for skeleton-based action recognition. It furnishes the 3D spatial coordinates of 25 joints for each human depicted in an action, as illustrated in Figure~\ref{fig:skeleton_examples} (a). For assessing the proposed methods, two evaluation protocols are suggested: Cross-Subject and Cross-View. The Cross-Subject setting involves 40,320 samples, with 16,560 allocated for training and evaluation, employing a split of 40 subjects into training and evaluation groups. In the case of Cross-View, comprising 37,920 and 18,960 samples, the evaluation uses camera 1 while training is conducted using cameras 2 and 3. Recently, an extended version of the original NTU-RGB+D dataset known as NTU-RGB+D 120 has been introduced. This extended dataset comprises 120 action classes and encompasses a total of 114,480 skeleton sequences, significantly expanding the scope. Additionally, the viewpoints have increased to 155. 

In Table~\ref{tab:table1} and Table~\ref{tab:table2}, we present the performance of recent skeleton-based techniques relevant to NTU-RGB + D and NTU-RGB + D 120 datasets, respectively. Note that in NTU-RGB+D, 'CS' stands for Cross-Subject, and 'CV' stands for Cross-View. For NTU-RGB + D120, there are two settings, \ie Cross-Subject (C-Subject), and Cross-Setup (C-Setup). 

\textcolor{black}{Based on the observation of the performance of these two datasets, we find that it's evident that existing algorithms have achieved impressive performances in the original NTU-RGB+D dataset. However, the newer NTU-RGB+D 120 poses a significant challenge, indicating that further advancements are needed to effectively address this more complex dataset. It's worth noting that the GCN-based methods achieved the leading results compared to the other two architectures. In addition to the very fundamental architectures (\ie RNNs, CNNs, and GCN), the most recent Transformer~\cite{vaswani2017attention} based methods also show their promising performance on both datasets. It's also easy to find that a hybrid Transformer and other architectures also further boost the overall performance of the 3D-SAR.}

%% file: sec/4_discussion.tex
\section{Discussion}
\label{sec:discussion}
\textcolor{black}{Considering the performance and attributes of the aforementioned deep architectures, several critical points warrant further discussion concerning the criteria for architecture selection. In terms of accuracy and robustness, GCNs demonstrate potential excellence by adeptly capturing spatial and temporal relationships among joints. RNNs exhibit proficiency in capturing temporal dynamics, while CNNs excel in identifying spatial features. When evaluating computational efficiency, CNNs boast faster processing capabilities owing to their parallel processing nature, contrasting with RNNs' slower sequential processing. Additionally, RNNs tend to excel in recognizing fine-grained actions, where temporal dependencies play a crucial role, while CNNs may better suit the recognition of gross motor actions based on spatial configurations. Considering factors like dataset size and hardware resources, the choice becomes more adaptable, contingent on the final model's scale. The size of the dataset and available computational resources for training become pivotal considerations, as different architectures might entail varying requirements.}  
\textcolor{black}{In summary, when recognizing actions reliant on temporal sequences, RNNs prove suitable for capturing the nuanced temporal dynamics within joint movements. In contrast, CNNs excel in identifying static spatial features and local patterns among joint positions. However, for comprehensive action recognition, leveraging both spatial and temporal relationships among joints, GCNs offer a beneficial approach when dealing with 3D skeletal data. }

\textcolor{black}{A possible in-practical solution can be also proposed to integrate not only one architecture but also a combination of them. This may make the final model absorb the advantages of each fundamental architecture. Furthermore, beyond the choice of deep architectures, the trajectory of 3D skeleton action recognition (SAR) navigation is a crucial consideration. Building upon our earlier discussions, we deduce that long-term action recognition, optimizing 3D-skeleton sequence representations, and achieving real-time operation remain significant open challenges. Moreover, annotating action labels for given 3D skeleton data remains exceptionally labor-intensive. Exploring avenues such as unsupervised or weakly-supervised strategies, along with zero-shot learning, may pave the way forward.}

%% file: sec/5_conclusion.tex
\section{Conclusion}
\label{sec:conclusion}
This paper presents an exploration of action recognition using 3D skeleton sequence data, employing four distinct neural network architectures. It underscores the concept of action recognition, highlights the advantages of skeleton data, and delves into the characteristics of various deep architectures. Unlike prior reviews, our study pioneers a data-driven approach, providing comprehensive insights into deep learning methodologies, encompassing the latest algorithms spanning RNN-based, CNN-based, GCN-based, and Transformer-based techniques. Specifically, our focus on RNN and CNN-based methods centers on addressing spatial-temporal information by leveraging skeleton data representations and intricately designed network architectures. In the case of GCN-based approaches, our emphasis lies in harnessing joint and bone correlations to their fullest extent. Furthermore, the burgeoning Transformer architecture has garnered significant attention, often employed in conjunction with other architectures for action recognition tasks.  Our analysis reveals that a fundamental challenge across diverse learning structures lies in effectively extracting pertinent information from 3D skeleton data. The topology graph emerges as the most intuitive representation of human skeleton joints, a notion substantiated by the performance metrics observed in datasets like NTU-RGB+D. However, this doesn't negate the suitability of CNN or RNN-based methods for this task. On the contrary, the introduction of innovative strategies, such as multi-task learning, shows promise for substantial improvements, particularly in cross-view or cross-subject evaluation protocols. Nevertheless, achieving further accuracy enhancements on datasets like NTU-RGB+D presents increasing difficulty due to the already high-performance levels attained. Hence, redirecting focus towards more challenging datasets, such as the enhanced NTU-RGB+D 120 dataset, or exploring other fine-grained human action datasets becomes imperative. Finally, we delve into an exhaustive discussion on the selection of foundational deep architectures and explore potential future pathways in 3D skeleton-based action recognition.

\section{Acknowledgements}
\label{sec:Acknowledgements}
This work was supported by National Natural Science Foundation of China (No. 62203476), Natural Science Foundation of Shenzhen (No. JCYJ20230807120801002). Data are available upon reasonable request.